\pdfoutput=1

\documentclass[11pt]{article}

\usepackage[]{EMNLP2023}

\usepackage{times}
\usepackage{latexsym}

\usepackage[T1]{fontenc}

\usepackage[utf8]{inputenc}

\usepackage{microtype}

\usepackage{inconsolata}

\usepackage{graphicx}

\usepackage{amsmath}
\usepackage{amssymb}

\usepackage{bbding}
\usepackage{multirow}
\usepackage{booktabs}
\usepackage{enumerate}
\usepackage{hyperref}

\title{CoT-MoTE: Exploring ConTextual Masked Auto-Encoder Pre-training with Mixture-of-Textual-Experts for Passage Retrieval}

\author{Guangyuan Ma\textsuperscript{\rm 1,2}\footnotemark[1], Xing Wu\textsuperscript{\rm 1,2,3}\footnotemark[1], Peng Wang\textsuperscript{\rm 1,2}, Songlin Hu\textsuperscript{\rm 1,2}\footnotemark[2] \\
  \textsuperscript{\rm 1}Institute of Information Engineering, Chinese Academy of Sciences, Beijing, China\\
  \textsuperscript{\rm 2}School of Cyber Security, University of Chinese Academy of Sciences, Beijing, China\\
  \textsuperscript{\rm 3}Kuaishou Technology, Beijing, China\\
  \texttt{\{maguangyuan,wuxing,wangpeng,husonglin\}@iie.ac.cn} \\}

\begin{document}
\maketitle

\renewcommand{\thefootnote}{\fnsymbol{footnote}} 
\footnotetext[1]{These authors contributed equally to this work.} 
\footnotetext[2]{Corresponding authors.} 
\renewcommand{\thefootnote}{\arabic{footnote}}

\begin{abstract}
Passage retrieval aims to retrieve relevant passages from large collections of the open-domain corpus.
Contextual Masked Auto-Encoding has been proven effective in representation bottleneck pre-training of a monolithic dual-encoder for passage retrieval. 
Siamese or fully separated dual-encoders are often adopted as basic retrieval architecture in the pre-training and fine-tuning stages for encoding queries and passages into their latent embedding spaces.
However, simply sharing or separating the parameters of the dual-encoder results in an imbalanced discrimination of the embedding spaces.
In this work, we propose to pre-train Contextual Masked Auto-Encoder with Mixture-of-Textual-Experts (CoT-MoTE). 
Specifically, we incorporate textual-specific experts for individually encoding the distinct properties of queries and passages.
Meanwhile, a shared self-attention layer is still kept for unified attention modelings.
Results on large-scale passage retrieval benchmarks show steady improvement in retrieval performances.
The quantitive analysis also shows a more balanced discrimination of the latent embedding spaces.
\end{abstract}

\section{Introduction}
Passage retrieval \cite{karpukhin-etal-2020-dense} aims to perform a relevance-based retrieval from large corpus collections with a given query. Pre-training with representation bottleneck \cite{lu-etal-2021-less, wang2022simlm} has been a hot topic in the effective pre-training of representation embeddings. Contextual Masked Auto-Encoding (CoT-MAE) \cite{wu2022contextual, wu2022query-as-context} is a typical representation bottleneck pre-training architecture, which utilizes an encoder-decoder structure and connects them merely with a single representation vector. The decoder is employed to reconstruct the whole contextual span, i.e. a nearby span or T5-generated pseudo-queries \cite{Rodrigo2019doc2query}, of a given passage. And the information of a contextual span will be compressed into a single representation vector, which gives good initialization of a monolithic dual-encoder for effective retrieval. 

\begin{figure}[t!]
\centering
\includegraphics[width=\linewidth]{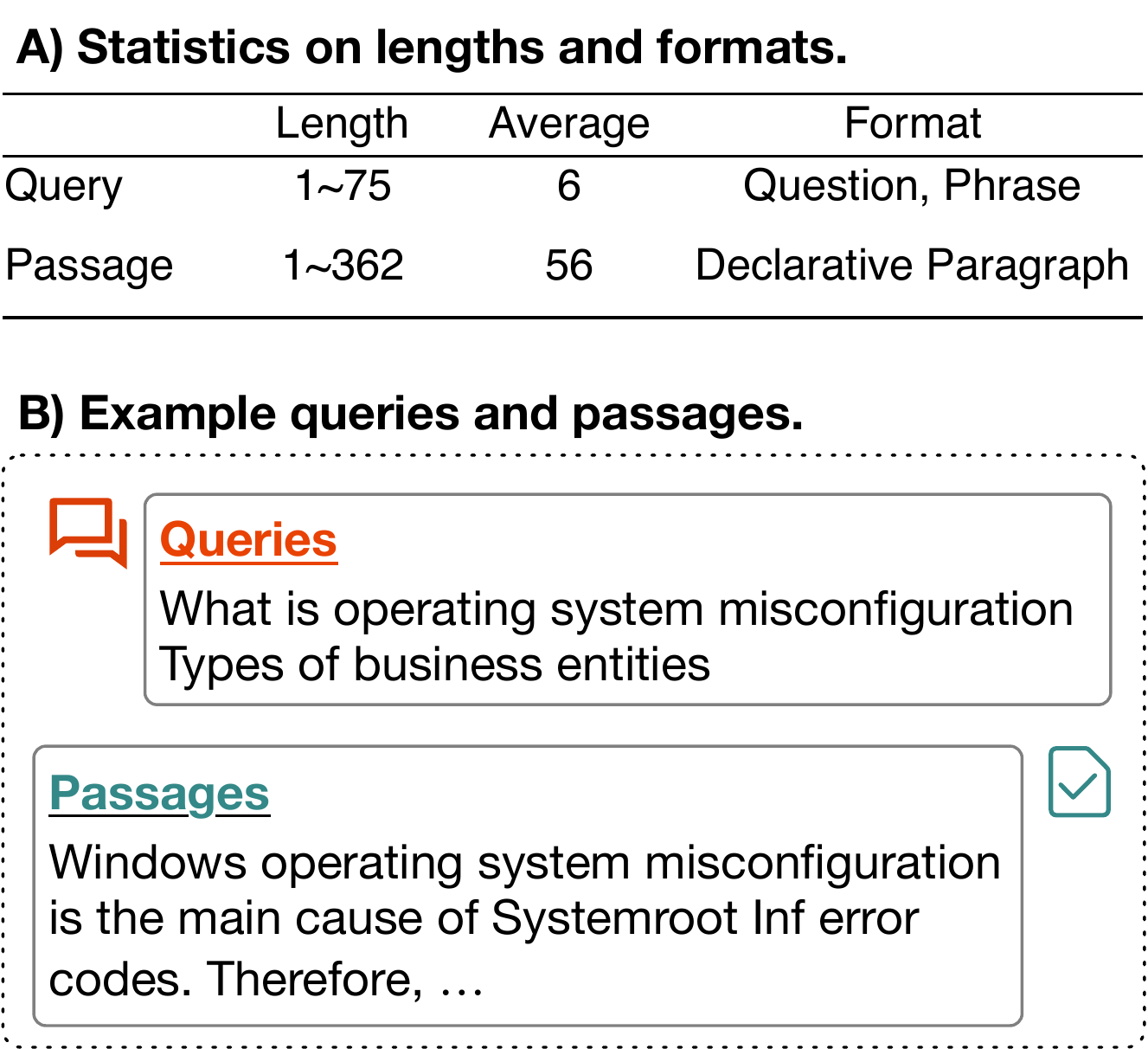}
\caption{
\textbf{A)} 
Statistics on the queries and passages of MS-MARCO corpus. 
The queries are often in question or phrase format, having 6 words on average. However, the passages are mostly declarative paragraphs which have 56 words on average.
\textbf{B)} 
Example queries and passages in MS-MARCO corpus.
}
\label{msmarco_qp_examples}
\end{figure}

Dual-encoder is a typical retrieval architecture used in passage retrieval. It often adopts Siamese or fully separated manners for encoding queries and passages into their latent embedding spaces. We statically analyze the average length and syntax of the query and passage sets (Figure \ref{msmarco_qp_examples}) in MS-MARCO \cite{tri2016msmarco} collections\footnote{\href{https://github.com/microsoft/MSMARCO-Passage-Ranking}{https://github.com/microsoft/MSMARCO-Passage-Ranking}}, a large real-world web search dataset. Statistics show significant differences in the traits of queries and passages. Queries are typically in short length and often presented in question or phrase format. On the contrary, passages are in long text format with declarative sentences. 

\paragraph{What issues lie in existing dual-encoders?}
Given the huge differences between queries and passages, the dual-encoder is expected to encode them into a discrete latent embedding space, for better modeling their unique traits. 
However, our preliminary studies in Section \ref{preliminary_study} with t-SNE visualization and performance measurement show that \textbf{a)} Siamese dual-encoder has better retrieval performances, but low discrimination with respect to query-passage embedding distribution. \textbf{b)} Fully separated dual-encoder encodes the embeddings with separation, but this hurts the retrieval performances. 
A similar observation was also discovered by \cite{dong-etal-2022-exploring}.
We also observe less mutual information of query-passage embeddings in fully separated dual-encoder, which verifies a discrete distribution that hinders the retrieval.

We believe the Siamese dual-encoder's covered distribution is not optimal for passage retrieval, given the differences between queries and passages. A more suitable embedding distribution should strike a balance between keeping distinct latent embedding spaces for modeling unique traits and allowing partial overlap for better query-passage interactions. This will promote mutual information between queries and passages.

\paragraph{Solution with MoTE.}
Feed-Forward-Network (FFN) preserves the most parameters of Transformers blocks. In the Multi-Modality domain, VLMo \cite{Wenhui2021VLMo} proposes to encode visual and textual signals with unique FFN experts for separately encoding different modalities. In our work, we propose to pre-train and fine-tune the Contextual Masked Auto-Encoder with Mixture-of-Textual-Experts (CoT-MoTE). CoT-MoTE is designed to capture different traits of queries and passages, while still encoding them in the same embedding space for a more balanced discrimination of the latent embedding spaces. 

As is shown in Figure \ref{CoT-MoTE_model}, here we use unified multi-head self-attention for modeling both queries and passages. But different Feed-Forward-Networks (FFN) are introduced individually for discrete semantic capturing. \textit{Query FFN expert (Q-FFN)} is designed to encode queries, which are typically short-length questions or phrases. while \textit{Passage FFN expert (P-FFN)} is designed to encode passages, which are typically long textual paragraphs. Different FFN experts discretely capture the unique traits of queries and passages, while a unified multi-head self-attention ensures joint modeling and enough mutual information with each other. 

To verify the effectiveness of our proposed methods, we conduct experiments on the large-scale web search benchmarks: MS-MARCO Passage Ranking \cite{tri2016msmarco}, TREC Deep Learning (DL) Track 2019 \cite{craswell2020overview} and 2020 \cite{craswell2021overview}. Experiments on large-scale passage retrieval benchmarks show steady improvement in retrieval performances. We also conduct several ablation studies to observe a better modeling ability with our methods.

Our contributions can be summarized as follows: \\
$\bullet$ We propose CoT-MoTE, Mixture-of-Textual-Experts for CoT-MAE, to effectively alleviate modeling issues of existing dual-encoders. \\
$\bullet$ We systematically analyze the performance boosts of CoT-MoTE with the retrieval baseline. \\
$\bullet$ Experiments show that CoT-MoTE achieves individual modeling abilities and balanced discrimination of queries and passages.

\section{Preliminary Studies}
\label{preliminary_study}
This section will introduce passage retrieval tasks with dual-encoder as preliminary knowledge. Then we will analyze existing issues existed in dual-encoders.

\subsection{Passage Retrieval with Dual-Encoders}
Given a large passage collection $\{p_1, p_2, ..., p_n\} \in P$, passage retrieval tasks aim to find relevant passages of a given query $q$ based on their similarities.

Dual-encoder is a typical retrieval architecture for passage retrieval. It often employs a PLM-based model as a backbone encoder for encoding representations. Formally, given a sequence of tokenized input queries and passages $\mathbb{T}$.

\begin{equation}
\mathbb{T} = \{CLS, t_1, ..., t_N, SEP\} \label{eq_1}
\end{equation}

The input texts are forwarded through L-layers ($l \in \{1, ..., L\}$) Transformers Blocks of PLM encoder. We denote the output hidden states as follows.

\begin{equation}
\mathbf{H}^{l} = \{\mathbf{h}_0^{l}, \mathbf{h}_1^{l}, ..., \mathbf{h}_N^{l}\}
\end{equation}

\begin{figure}[!ht]
\centering
\includegraphics[width=8cm]{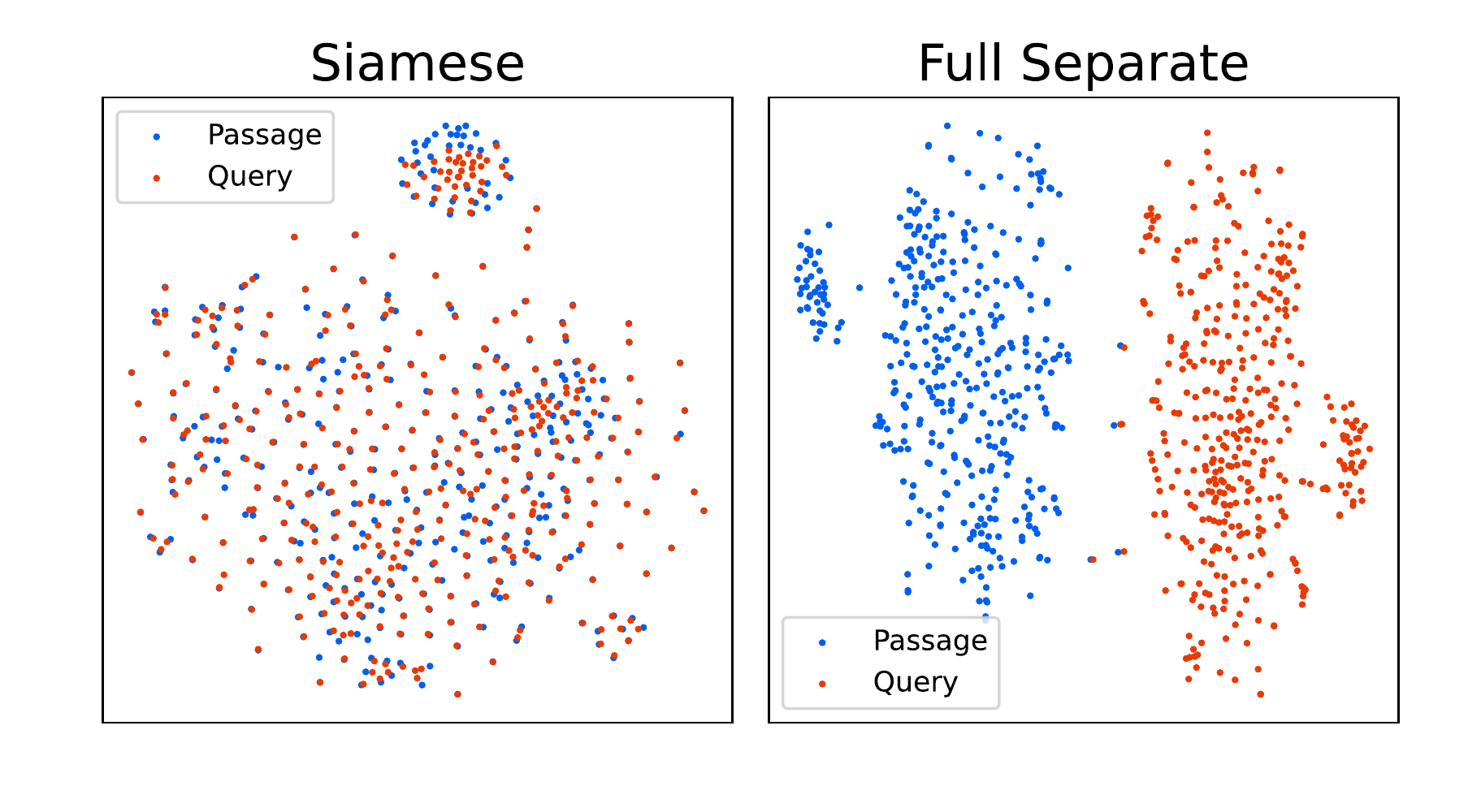}
\caption{
T-SNE visualization of the embeddings of MS-MARCO dev set generated by Siamese and Fully separated dual-encoder of CoT-MAE.
}
\label{cotmae_prestudy_tsne}
\end{figure}

Hidden states at [CLS] positions of the last layers are commonly used as the dense representations of dual-encoders. Dense Representations model the semantics at the sentence level.

\begin{equation}
\mathbf{DEN} = \mathbf{h}_0^{last}
\end{equation}

Existing works, eg. SPLADE \cite{Thibault2021splade, Thibault2021splade_v2} and LexMAE \cite{Tao2022lexmae}, also focus on learning PLM-based dual-encoders for capturing bag-of-words semantic sparse representations. Sparse representations are focusing on encoding semantics at the token level. The hidden states of the last layer $\mathbf{h}^{last}$ are firstly projected to vocab spaces with the transposed embedding matrix $\mathbf{E^\top}$. Then ReLU and log saturation operations are performed to obtain the sparse representations.

\begin{equation}
\mathbf{SPR} = \max_{i \in \{1, ..., N\}} \log{(1 + ReLU(\mathbf{h}_i^{last}\mathbf{E^\top}))}
\end{equation}

\noindent here $i \in \{1, ..., N\}$ denotes the sequence length spaces. ReLU keeps non-negative hidden states as token frequencies produced by PLM. And log saturation suppresses the dominant frequencies for keeping sparsity. The above operations are common practices for pooling sparse representations.

\paragraph{Siamese or Fully Separated Dual-encoders}
Dual-encoder is composed of a query encoder and a passage encoder. A common practice for passage retrieval employs a Siamese dual-encoder, which shares the parameters between the query encoder and passage encoder. Meanwhile, fully separated dual-encoders are also applied to encode the queries and passages with encoders that do not share any parameters.

\begin{table}[!ht]
    \centering
    \resizebox{\linewidth}{!}{
    \begin{tabular}{l|c|cc}
    \toprule 
        ~ & MRR@10 & KL & JS \\ 
        \midrule
        Siamese & 39.4 &  7.24e-5 & 1.81e-5 \\
        Fully Separate & 39.0 & 1.42e-4 & 3.54e-5 \\
    \bottomrule
    \end{tabular}
    }
    \caption{
    Performances and Average of KL \& JS scores for Siamese and Fully separated CoT-MAE dual-encoders.
    }
    \label{table_kl_js}
\end{table}

\paragraph{Retrieval based on Similarities}
Dot product or cosine similarity is often used as the measurement of query-passage similarities $s(q, p)$. Here we take the dot product as an example, 

\begin{equation}
s_{den}(q, p) = \mathbf{DEN}_q \cdot \mathbf{DEN}_p
\end{equation}

\begin{equation}
s_{spr}(q, p) = \mathbf{SPR}_q \cdot \mathbf{SPR}_p
\end{equation}

Large-scale similarity search is performed within passage collections $P$ based on dense retrieval $s_{den}(q, p)$, sparse retrieval $s_{spr}(q, p)$ or hybrid retrieval $s_{den}(q, p) + s_{spr}(q, p)$ distances.

\subsection{Discrimination Issue in Dual-Encoders}
As discussed in the introduction section, the dual-encoder is expected to encode the query and passage embeddings into a discrete latent space for better modeling their own semantics. To dive deep into the distribution of dual-encoder embeddings, we randomly encode 400 query-passage pairs from the MS-MARCO dev corpus with CoT-MAE dual-encoders. Then we project the embeddings to 2-dimensional space with t-SNE visualization. The cosine similarity is used as the distance measure.

As is shown in Figure \ref{cotmae_prestudy_tsne}, in Siamese encoders, the embeddings are sharing the same latent space. This schema makes low discrimination for capturing the unique traits of queries and passages. On the contrary, the fully separated encoders tend to encode the embeddings into a separated space, making low mutual interaction with queries and passages.

\paragraph{Performances of Dual-encoders}
We compare the performance of Siamese and fully separated dual-encoders using the fine-tuning pipeline in CoT-MAE and report the results in Table \ref{table_kl_js}. To assess the mutual information between queries and passages, we use KL divergence and JS divergence to measure the similarities in their distributions. Appendix \ref{appendix_kl_js_div} provides details on the calculations.

While T-SNE visualization demonstrates effective embedding discrimination between queries and passages for fully separated dual-encoder, Table \ref{table_kl_js} reveals inferior MRR@10 results on MS-MARCO dev sets. Our evaluation of KL and JS divergence indicates that the fully separated dual-encoder has higher divergences, implying less interaction between query and passage embeddings. In contrast, the Siamese dual-encoder exhibits higher retrieval performance and lower query-passage divergence.

However, we argue that the embeddings generated by Siamese models may not be the best fit for capturing the unique characteristics of queries and passages, given their distinctive differences in terms of length and syntax. We believe the individual modeling of queries and passages with individual FFN experts will be a simple yet effective solution to alleviate the problem within Siamese and fully separated dual-encoder.

\begin{figure*}[!htbp]
\centering
\includegraphics[width=\linewidth]{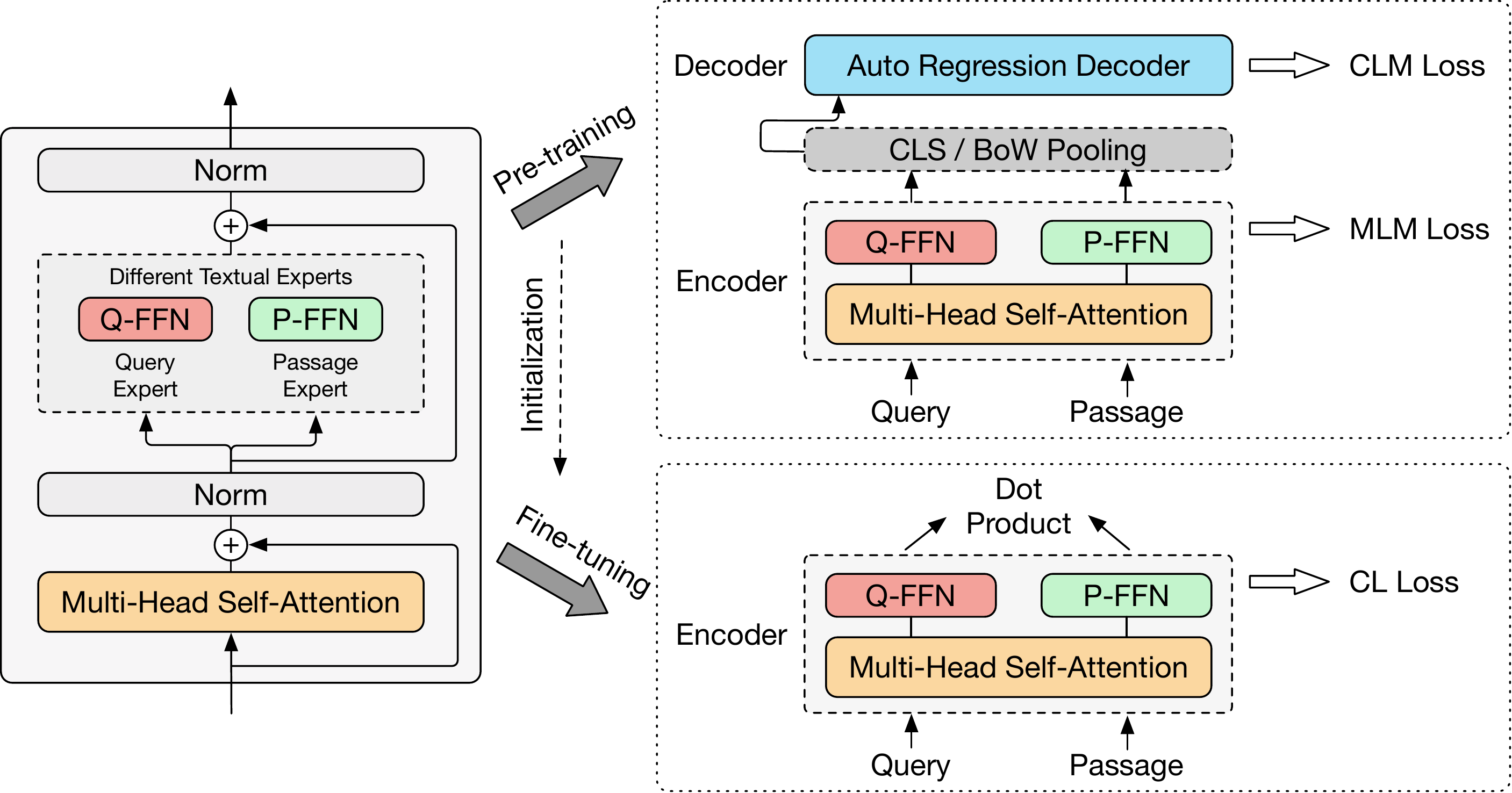}
\caption{
The implementation of CoT-MoTE. Query Expert (Q-FFN) and Passage Expert (P-FFN) are introduced in Transformers blocks to individually encode queries and passages in the pre-training and fine-tuning stages. 
}
\label{CoT-MoTE_model}
\end{figure*}

\section{Mixture-of-Textual-Experts Pre-training}
To counter the discrimination problems within existing dual-encoders, in our work, we propose to pre-train the queries and passages with specific textual experts. we plot our implementation in Figure \ref{CoT-MoTE_model}.

\subsection{Mixture-of-Textual-Experts}
We adapt on Transformers Blocks for better modeling textual traits. Formally, in $l$-layer Transformers blocks, it receives hidden states $h_{l-1}$ from the $l-1$ layer as inputs.

\paragraph{Unified Self-Attention}
To counter the discrete distribution of the fully separated dual-encoder, we preserve a unified self-attention ($Attn$) for joint modeling of the attention mechanism of both query and passage textual features like normal Transformers. This helps to learn unified embedding distribution.

\begin{equation}
Attn_{out} = LN(Attn(h_{l-1}) + h_{l-1})
\end{equation}

\noindent where $Attn_{out}$ is the concatenated attention outputs of unified multi-head self-attention. LN means layer norm in Transformers.

\paragraph{Indivisual Textual Experts}
\label{indivisual_textual_experts}
We introduce separated Feed-Forward-Networks, Q-FFN, and P-FFN, for individual query and passage experts. Q-FFN receives queries as inputs, which focus on capturing traits of short questions or phrases. P-FFN receives passages as inputs, which focus on long declarative sentence modeling. This helps to learn different embeddings with enough discrimination. Suppose the $Attn_{out}$ is composed of $Attn_{q}$ and $Attn_{p}$ based on input text types. The output hidden states of the current layer are as follows.

\begin{equation}
h_l^{q} = LN(Q-FFN(Attn_{q}) + Attn_{q})
\end{equation}

\begin{equation}
h_l^{p} = LN(P-FFN(Attn_{p}) + Attn_{p})
\end{equation}

\noindent where $h_l^{q}$ means $l$-layer query hidden states, $h_l^{p}$ means $l$-layer passage hidden states.

\subsection{Pre-training}
Pre-training with representation bottleneck \cite{lu-etal-2021-less, wang2022simlm} has been a hot topic for effective retrieval representation learning. In our work, we choose to pre-train our Mixture-of-Textual-Experts based on a popular passage retrieval pre-training schema, Contextual Masked Auto-Encoding (CoT-MAE). CoT-MAE \cite{wu2022contextual, wu2022query-as-context} is a typical representation bottleneck pre-training architecture. It composes a 12-layer BERT encoder and a single-layer shallow decoder in the pre-training stages. CoT-MAE utilizes an encoder-decoder structure and connects them merely with a single representation vector.

Formally, given the input text $\mathbb{T}$ in Equation \ref{eq_1}, we randomly mask a certain percentage of tokens in $\mathbb{T}$ with a specified mask token, e.g., $[M]$.

\begin{equation}
mask(\mathbb{T}) = \{CLS, t_1, M, t_3, ..., t_N, SEP\}
\end{equation}

Then the masked inputs are forwarded through the encoder ($Enc$). As discussed in the above subsection, different textual experts are used based on input text types.
Cross-Entropy loss is used to reconstruct the tokens at masked positions. This loss is also denoted as the Masked Language Modeling (MLM) objective.

\begin{equation}
\mathcal{L}_{enc}=-\sum_{i \in mask(\mathbb{T})} \log p(t_i|Enc(mask(\mathbb{T})))
\end{equation}

A single-layer decoder receives the pooled dense or sparse representations as inputs. Given the representations from the encoder as $DEN$ and $SPR$, an auto-regression decoder ($Dec$) is applied to reconstruct the \textbf{whole contextual span} $\mathbb{T}_{Dec}$, i.e. a nearby span or T5-generated pseudo-queries \cite{Rodrigo2019doc2query}, of a given passage. Following the settings in \cite{wu2022query-as-context}, here we choose pseudo-queries as the decoder inputs for simplicity.
The decoder utilizes Casual Language Model (CLM) as generative signals, whose loss is formulated as

\begin{equation}
\mathcal{L}_{dec}=-\sum_{i \in \mathbb{T}_{Dec}} \log p(t_{i+1}|Dec(\mathbb{T}_{Dec}))
\end{equation}

In the fine-tuning and inferencing stages, the decoder will be discarded. The optimized encoder is used to initialize the dual encoder for effective passage retrieval.

\begin{table*}[!ht]
\centering
\resizebox{\linewidth}{!}{
    \begin{tabular}{l|c|ccc|c|c}
    \toprule  
    ~ & \textbf{Mined} & \multicolumn{3}{c|}{\textbf{MS-MARCO}} & \textbf{TREC DL 19} & \textbf{TREC DL 20}\\
    \textbf{Model} & \textbf{HN} & MRR@10 & R@50 & R@1k & nDCG@10 & nDCG@10 \\
    \midrule
    BM25 \cite{robertson2009probabilistic} & ~ &  18.7  & 59.2  & 85.7  & 51.2 & 47.7 \\
    SPLADE \cite{Thibault2021splade} & ~ & 32.2  & - & 95.5  & 66.5 & - \\
    SEED \cite{lu-etal-2021-less} & ~ & 33.9  & - & 96.1  & - & - \\
    RocketQA \cite{qu-etal-2021-rocketqa} & $\checkmark$ & 37.0  & 85.5  & 97.9  & - & - \\
    coCondenser \cite{gao-callan-2022-unsupervised} & $\checkmark$  & 38.2  & 86.5  & 98.4  & 71.7 & 68.4 \\
    SimLM \cite{wang2022simlm} & $\checkmark$ & 39.1 & - & 98.6 & - & - \\
    CoT-MAE \cite{wu2022contextual} & $\checkmark$ & 39.4 & 87.0 & 98.7 & 70.9 & 70.4 \\
    LED \cite{zhang2022LED} & $\checkmark$ & 39.6 & 86.6 & 98.3 & 70.5 & 67.9 \\
    MASTER \cite{zhou2022master} & $\checkmark$ & 40.4 & - & \textbf{98.8} & - & - \\
    LexMAE \cite{Tao2022lexmae} & $\checkmark$ & 40.8  & - & 98.5  & - & - \\
    CoT-MAE v2 \cite{wu2023cotmaev2} & $\checkmark$ & 41.4 & 89.4 & 98.7 & 74.5 & 70.8 \\
    \midrule
    CoT-MoTE & $\checkmark$ & \textbf{42.0}\textsuperscript{**} & \textbf{89.7}\textsuperscript{*} & 98.7 & \textbf{75.5}\textsuperscript{**} & \textbf{72.9}\textsuperscript{**} \\
    \bottomrule
    \end{tabular}
}
\caption{
Main results on MS-MARCO passage ranking and TREC DL datasets. The best score is marked in bold. Checkmarks for Mined HN represent whether the retriever is trained on mined hard negatives. Two-tailed t-tests demonstrate statistically significant improvements of CoT-MoTE over baselines ( $* \leq $ 0.05, $** \leq $ 0.01 ).
}
\label{table_results_main}
\end{table*}

\subsection{Fine-tuning and Inference}
In fine-tuning and inference stages, we share the parameters of unified self-attentions across query and passage encoders. But we use individual pre-trained textual experts in different encoders as discussed in Section \ref{indivisual_textual_experts}.

\paragraph{Fine-tuning} We initialize the dual encoders from the pre-trained encoder and fine-tune them on large-scale benchmarks to verify the effectiveness of our proposed Mixture-of-Textual-Experts (MoTE). Following \cite{wu2023cotmaev2}, we use Contrastive Learning Loss (CL Loss) to train our retrievers with hard negatives.

\begin{equation}
\mathcal{L}_{\mathrm{ft}}=-\log \frac{\exp({s(q, p^{+})}))}{\sum \exp(({s(q, p^{+})} + {s(q, p^{-})}))}
\end{equation}

\noindent where $s(q, p)$ is hybrid retrieval similarities $s_{den}(q, p) + s_{spr}(q, p)$. We will perform detailed comparisons of dense, sparse, and hybrid performances with MoTE in our experiments. ${p^{+}}$ is the positive passages labeled by human annotators. ${p^{-}}$ is the hard negative sets provided via BM25 or dense retrieval. 

Following typical passage retrieval fine-tuning pipelines \cite{wu2023cotmaev2}, we first train our dual-encoders with BM25 negatives provided by the MS-MARCO training set. And then we use retriever-s1 to mine effective hard negatives for fine-tuning the retriever-s2. We directly reuse the above pipelines from CoT-MAE without any hypermeter tuning.

\paragraph{Inference}
Following \cite{wu2023cotmaev2}, we get the CLS dense vectors and Top-${k}$ sparse vectors for similarity computation. Faiss \cite{johnson2019billion} is used for dense vector search, while PyTorch Sparse matrix multiplication is used for sparse vector retrieval.

\section{Experiments}
In this section, we first introduce the experiment settings of the pre-training and fine-tuning stages. Then we report the main results of our proposed methods.

\subsection{Pre-training}
The official checkpoint of BERT-base-uncased is used to initialize the pre-training of CoT-MoTE. Both Q-FFN and P-FFN are initialized from the original FFN weights. We directly reuse the off-the-shelf pre-training corpus, MS-MARCO documents with 3.2M docs, used in CoT-MAE-qc \cite{wu2022query-as-context} for reproducibility. The corpus was cut to a max length of 144. DocT5query \cite{Rodrigo2019doc2query} was used to generate 5 pseudo-queries for each passage spans. We randomly sample the query-passage pairs and feed them into the CoT-MoTE encoder. The decoder takes encoder representation vectors and contextual spans as inputs. Following CoT-MAE \cite{wu2022query-as-context}, we use queries as the contextual spans for the inputs of the CoT-MoTE decoder. The model is pre-trained using the AdamW optimizer for a maximum of 80k steps, with a learning rate of 3e-4, batch size of 2k, and a linear scheduler with a warmup ratio of 0.1. The whole process is performed on 8 Tesla A100 for 12 hours. The seed is set to 42 in our analysis for reproducibility.

\subsection{Fine-tuning}
We fine-tune on MS-MARCO Passage Ranking \cite{tri2016msmarco} and TREC Deep Learning (DL) Track 2019 \cite{craswell2020overview} and 2020 \cite{craswell2021overview} to evaluate the retrieval performances of our methods. We directly reuse the fine-tuning settings in \cite{wu2023cotmaev2}. The model is fine-tuned using the AdamW optimizer for 3 epochs, with a learning rate of 2e-4, and a batch size of 64. The number of negative passages is 15 and the negative depth is 200. 

\subsection{Baselines}
We compare our method with multiple sparse, dense, and hybrid retrieval baselines. Sparse retrieval baselines include BM25 \cite{robertson2009probabilistic}, SPLADE \cite{Thibault2021splade} and LexMAE \cite{Tao2022lexmae}. Dense retrieval baselines include SEED \cite{lu-etal-2021-less}, RocketQA \cite{qu-etal-2021-rocketqa}, coCondenser \cite{gao-callan-2022-unsupervised}, SimLM \cite{wang2022simlm}, CoT-MAE \cite{wu2022contextual}, LED \cite{zhang2022LED}, and MASTER \cite{zhou2022master}. Hybrid retrieval baselines include CoT-MAE v2 \cite{wu2023cotmaev2}. We take the non-distilled results from their papers as baselines. Because distillation from a strong re-ranker is out of the scope in our paper and we leave this to our future works.

\subsection{Main Results}
\label{subsection_main_results}
Our work achieves state-of-the-art results on multiple baselines in Table \ref{table_results_main}. Compared to the latest hybrid retrieval baseline CoT-MAE v2, which integrates the multi-view pre-training technique, our method outperforms it by +0.6 MRR@10, +0.3 R@50 on the MS-MARCO passage ranking benchmark, and +1.0, +2.1 nDCG@10 on TREC-DL 2019 \& 2020 benchmarks. 
Note that our work also shares the same retrieval speed with CoT-MAE v2, because integrating Mix-of-Textual-Experts in the pre-training and fine-tuning stages does not require a deeper Transformers architecture or additional retrieval pipelines.

\begin{table}[!tbp]
\centering
\resizebox{\linewidth}{!}{
    \begin{tabular}{l|cccc}
    \toprule  
    \textbf{Model} & MRR@10 & R@1 & R@50 & R@1k \\
    \midrule
    \multicolumn{5}{l}{\textbf{Dense Retrieval}} \\
    \midrule
    Siamese & 39.3 & 26.1 & \textbf{88.4} & \textbf{98.8} \\
    Full sep & 38.4 & 25.3 & 87.3 & 98.5 \\
    \textbf{CoT-MoTE} & \textbf{39.7} & \textbf{26.7} & 88.1 & 98.7 \\
    
    \midrule
    \multicolumn{5}{l}{\textbf{Sparse Retrieval}} \\
    \midrule
    Siamese & 39.3 & 26.2 & 86.6 & 98.4 \\
    Full sep & 39.4 & 26.4 & 86.8 & 98.4 \\
    \textbf{CoT-MoTE} & \textbf{40.5} & \textbf{27.1} & \textbf{88.4} & \textbf{98.8} \\
    
    \midrule
    \multicolumn{5}{l}{\textbf{Hybrid Retrieval}} \\
    \midrule
    Siamese & 40.9 & 27.3 & 89.2 & 98.8 \\
    Full sep & 40.5 & 27.1 & 89.4 & 98.5 \\
    \textbf{CoT-MoTE} & \textbf{42.0} & \textbf{28.6} & \textbf{89.7} & 98.7 \\
    
    \bottomrule
    \end{tabular}
}
\caption{Ablation study for dense, sparse, and hybrid retrieval performances with Siamese, fully separated, and CoT-MoTE. 
}
\label{table_ablation_mote_siamese_full_sep}
\end{table}

\section{Analysis}
In this section, we provide detailed analyses (\textbf{A}) for below scientific questions (\textbf{Q}). \\
\textbf{Q1.} How does MoTE contribute to in-domain retrieval performances compared to previous dual-encoders? \\
\textbf{Q2.} Does MoTE serves as the unique expert for individual modeling queries and passages? \\
\textbf{Q3.} How does MoTE improve the query-passage distribution in latent embedding spaces? \\
\textbf{Q4.} How does the mutual information of query-passage embeddings in MoTE compare to previous dual-encoders?

\subsection{Comparision of Different Dual-encoders}
To fully explore the performance differences of CoT-MoTE with Siamese and fully separate architectures, we pre-train and fine-tune the above architectures with the following settings: 1) Siamese dual-encoder is pre-trained and fine-tuned by removing qry-FFN from CoT-MoTE. 2) Fully separated dual-encoder is pre-trained and fine-tuned with completely separated encoders. We keep other pipelines the same as our main result settings.

\begin{table}[!tbp]
\centering
\resizebox{\linewidth}{!}{
    \begin{tabular}{l|cccc}
    \toprule  
    ~ & MRR@10 & R@1 & R@50 & R@1k \\
    \midrule
    \multicolumn{5}{l}{\textbf{CoT-MoTE}} \\
    \midrule
    Dense & 39.7 & 26.7 & 88.1 & 98.7 \\
    Sparse & 40.5 & 27.1 & 88.4 & 98.8 \\
    Hybrid & 42.0 & 28.6 & 89.7 & 98.7 \\

    \midrule
    \multicolumn{5}{l}{\textbf{Switch Pre-training Corpus of FFN-Experts }} \\
    \midrule
    Dense & 37.9 (\textcolor{red}{-1.8}) & 25.3 (\textcolor{red}{-1.4}) & 86.2 (\textcolor{red}{-1.9}) & 98.5 (\textcolor{red}{-0.2}) \\
    Sparse & 37.7 (\textcolor{red}{-2.8}) & 25.1 (\textcolor{red}{-2.0}) & 85.5 (\textcolor{red}{-3.3}) & 98.1 (\textcolor{red}{-0.7}) \\
    Hybrid & 40.2 (\textcolor{red}{-1.8}) & 26.9 (\textcolor{red}{-1.7}) & 88.3 (\textcolor{red}{-1.4}) & 98.5 (\textcolor{red}{-0.2})  \\
    
    \bottomrule
    \end{tabular}
}
\caption{Ablation study for dense, sparse, and hybrid retrieval performances when switching the pre-training corpus for qry-FFN and psg-FFN of CoT-MoTE. 
}
\label{table_ablation_switch_pretrain_corpus}
\end{table}

\paragraph{A1.} We compare the dense, sparse, and hybrid retrieval results of CoT-MoTE with Siamese and fully separate dual-encoders in Table \ref{table_ablation_mote_siamese_full_sep}. Results show that 1) Fully separated dual-encoders achieve lower performances than the Siamese encoder. 2) CoT-MoTE achieves a significant boost on all retrieval pipelines compared with other dual-encoders. This verifies that our proposed MoTE significantly contributes to the retrieval baseline with respect to traditional Siamese and fully separated dual-encoders. 

Please note that CoT-MoTE has the same retrieval speed as other dual-encoders, as discussed in Section \ref{subsection_main_results}. Despite using more model parameters, we believe that the additional parameter size we used is not the decisive factor for better retrieval in our scenario. Individual textual modeling with MoTE and joint attention should be the key to the performance increase. In contrast, the fully separated dual-encoder employs more parameters than CoT-MoTE and Siamese encoders but performs the worst.

\subsection{Effect of Switching Pre-training Corpus of FFN-Experts}
To further verify that CoT-MoTE models unique traits of queries and passages with individual FFN experts, we switch the pre-training corpus for CoT-MoTE encoders, i.e. queries for psg-FFN and passages for qry-FFN. The rest of the settings for pre-training and fine-tuning is kept unchanged.

\paragraph{A2.} Results from Table \ref{table_ablation_switch_pretrain_corpus} show that switching the pre-training corpus for FFN Experts of CoT-MoTE significantly damages the retrieval performances. This verifies that FFN experts serve as individual workers for capturing the unique features of queries and passages.

\begin{figure*}[!htbp]
\centering
\includegraphics[width=\linewidth]{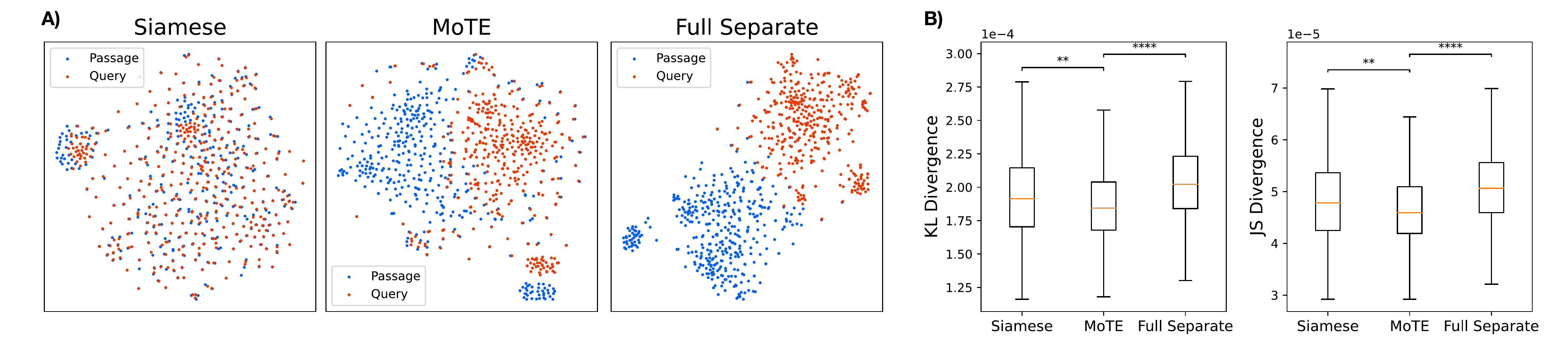}
\caption{
T-SNE visualization (A) and KL, JS divergences (B) with different dual-encoders. Two-tailed t-tests demonstrate statistically significant improvements of CoT-MoTE over baselines ( $** \leq $ 0.01, $**** \leq $ 0.0001 ).
}
\label{CoT-MoTE_tsne_kl_js}
\end{figure*}

\subsection{Quantitive Analysis for Embeddings}
To analyze the embedding distributions and query-passage mutual information of our proposed method, we plot and compare the t-SNE visualization results and KL, JS divergences of Siamese, MoTE, and Fully Separated dual encoders.

\paragraph{A3.} Figure \ref{CoT-MoTE_tsne_kl_js} (A) illustrates that MoTE exhibits local clustering while still partially covering each other. This results in a well-balanced embedding distribution that incorporates both local discrimination and global interactivity, which we believe contributes to its superior retrieval performance.

\paragraph{A4.} We also compare the KL and JS divergences of different dual-encoders in Figure \ref{CoT-MoTE_tsne_kl_js} (B). MoTE achieves statistically lower query-passage divergences, indicating that our method preserves the ability to model discretely while maintaining or even increasing the mutual information between queries and passages. This verifies that our approach does not compromise the interaction between queries and passages.

\section{Related Works}
With the development of large-scale pre-training technologies, PLM-based retrieval, i.e. BERT-based \cite{devlin-etal-2019-bert} dual-encoders, has been popular among the IR community \cite{karpukhin-etal-2020-dense, qu-etal-2021-rocketqa, Thibault2021splade}. 

\paragraph{Pre-training with Representation Bottleneck} Pre-training on PLMs with a representation bottleneck has gained popularity as an effective way to enhance their initialization of dual encoders. This involves an encoder-decoder structure connected by a single representation vector, which is improved through pre-tasks of the decoder.
Dense retrieval methods often rely on CLS or averaged pooled vectors to improve retrieval, including SEED \cite{lu-etal-2021-less}, SimLM \cite{wang2022simlm}, RetroMAE \cite{xiao-etal-2022-retromae}.
LexMAE \cite{Tao2022lexmae} employs an encoder-decoder structure to enhance the sparse vector through a sparse representation bottleneck. Recent studies have also focused on learning better PLMs with hybrid retrieval. The CoT-MAE series \cite{wu2022contextual, wu2022query-as-context} learns a better encoder for dense \cite{wu2022contextual}, sparse, and hybrid \cite{wu2023cotmaev2} with contextual span information, and also uses the representation bottleneck structure. The decoder in CoT-MAE is used to reconstruct the entire contextual span, such as a nearby span or T5-generated pseudo-queries \cite{Rodrigo2019doc2query}, of any given passage. Our Mix-of-Textual-Experts is based on CoT-MAE as it was the previous state-of-the-art method, and it can be easily reproduced with an open-source code base and training corpus.

\paragraph{Indivisual FFN-Experts} Feed Forward Network (FFN) preserves the most parameters and is treated as unique experts in Mix-of-Expert structures. Instead of employing a gate mechanism and routing the output of each expert, we directly assign the queries and passages to two different FFNs. Because we aim to counter the discrimination issue in embedding distribution, separated but not mixed FFN experts are essential for individually capturing unique textual traits. In the Multi-Modality domain, VLMo \cite{Wenhui2021VLMo} and BEiT-3 \cite{wang2022beit3} proposes separated FFN experts for modeling different modalities. In contrast, we introduce individual FFN experts to capture different traits of textual information in queries and passages, so as to counter the discrimination issue in the embedding of dual encoders.

\section{Conclusions}
This paper proposes to pre-train with the Mixture-of-Textual-Experts to counter the imbalanced discrimination issue in existing dual encoders. Textual-specific experts are introduced for individual modeling of the distinct traits of queries and passages. Results on large-scale web benchmarks show steady improvement in retrieval performances. Quantitive analysis shows a more balanced distribution of query-passage embeddings for dual-encoders.

\section*{Limitations}
The pre-training of CoT-MoTE requires additional pseudo-queries generated by docT5query or other query-generation models. Although we directly use the off-of-shelf CoT-MAE pre-training corpus for our reproduction, we argue that this process increases the GPU resource usage for generalized pre-training. We will further incorporate novel methods, e.g. curriculum learning, to alleviate this issue. We leave this to our future work.

\section*{Ethics Statement}
All authors claim to comply with the \href{https://www.aclweb.org/portal/content/acl-code-ethics}{ACL Ethics Policy}.

\bibliography{anthology,custom}

\begin{thebibliography}{25}
\expandafter\ifx\csname natexlab\endcsname\relax\def\natexlab#1{#1}\fi

\bibitem[{Craswell et~al.(2021)Craswell, Mitra, Yilmaz, and
  Campos}]{craswell2021overview}
Nick Craswell, Bhaskar Mitra, Emine Yilmaz, and Daniel Campos. 2021.
\newblock \href {http://arxiv.org/abs/2102.07662} {Overview of the trec 2020
  deep learning track}.

\bibitem[{Craswell et~al.(2020)Craswell, Mitra, Yilmaz, Campos, and
  Voorhees}]{craswell2020overview}
Nick Craswell, Bhaskar Mitra, Emine Yilmaz, Daniel Campos, and Ellen~M.
  Voorhees. 2020.
\newblock \href {http://arxiv.org/abs/2003.07820} {Overview of the trec 2019
  deep learning track}.

\bibitem[{Devlin et~al.(2019)Devlin, Chang, Lee, and
  Toutanova}]{devlin-etal-2019-bert}
Jacob Devlin, Ming-Wei Chang, Kenton Lee, and Kristina Toutanova. 2019.
\newblock \href {https://doi.org/10.18653/v1/N19-1423} {{BERT}: Pre-training of
  deep bidirectional transformers for language understanding}.
\newblock In \emph{Proceedings of the 2019 Conference of the North {A}merican
  Chapter of the Association for Computational Linguistics: Human Language
  Technologies, Volume 1 (Long and Short Papers)}, pages 4171--4186,
  Minneapolis, Minnesota. Association for Computational Linguistics.

\bibitem[{Dong et~al.(2022)Dong, Ni, Bikel, Alfonseca, Wang, Qu, and
  Zitouni}]{dong-etal-2022-exploring}
Zhe Dong, Jianmo Ni, Dan Bikel, Enrique Alfonseca, Yuan Wang, Chen Qu, and Imed
  Zitouni. 2022.
\newblock \href {https://aclanthology.org/2022.emnlp-main.640} {Exploring dual
  encoder architectures for question answering}.
\newblock In \emph{Proceedings of the 2022 Conference on Empirical Methods in
  Natural Language Processing}, pages 9414--9419, Abu Dhabi, United Arab
  Emirates. Association for Computational Linguistics.

\bibitem[{Formal et~al.(2021{\natexlab{a}})Formal, Lassance, Piwowarski, and
  Clinchant}]{Thibault2021splade_v2}
Thibault Formal, Carlos Lassance, Benjamin Piwowarski, and St{\'{e}}phane
  Clinchant. 2021{\natexlab{a}}.
\newblock \href {http://arxiv.org/abs/2109.10086} {{SPLADE} v2: Sparse lexical
  and expansion model for information retrieval}.
\newblock \emph{CoRR}, abs/2109.10086.

\bibitem[{Formal et~al.(2021{\natexlab{b}})Formal, Piwowarski, and
  Clinchant}]{Thibault2021splade}
Thibault Formal, Benjamin Piwowarski, and St{\'{e}}phane Clinchant.
  2021{\natexlab{b}}.
\newblock \href {https://doi.org/10.1145/3404835.3463098} {{SPLADE:} sparse
  lexical and expansion model for first stage ranking}.
\newblock In \emph{{SIGIR} '21: The 44th International {ACM} {SIGIR} Conference
  on Research and Development in Information Retrieval, Virtual Event, Canada,
  July 11-15, 2021}, pages 2288--2292. {ACM}.

\bibitem[{Gao and Callan(2022)}]{gao-callan-2022-unsupervised}
Luyu Gao and Jamie Callan. 2022.
\newblock \href {https://doi.org/10.18653/v1/2022.acl-long.203} {Unsupervised
  corpus aware language model pre-training for dense passage retrieval}.
\newblock In \emph{Proceedings of the 60th Annual Meeting of the Association
  for Computational Linguistics (Volume 1: Long Papers)}, pages 2843--2853,
  Dublin, Ireland. Association for Computational Linguistics.

\bibitem[{Johnson et~al.(2019)Johnson, Douze, and
  J{\'e}gou}]{johnson2019billion}
Jeff Johnson, Matthijs Douze, and Herv{\'e} J{\'e}gou. 2019.
\newblock Billion-scale similarity search with {GPUs}.
\newblock \emph{IEEE Transactions on Big Data}, 7(3):535--547.

\bibitem[{Karpukhin et~al.(2020)Karpukhin, Oguz, Min, Lewis, Wu, Edunov, Chen,
  and Yih}]{karpukhin-etal-2020-dense}
Vladimir Karpukhin, Barlas Oguz, Sewon Min, Patrick Lewis, Ledell Wu, Sergey
  Edunov, Danqi Chen, and Wen-tau Yih. 2020.
\newblock \href {https://doi.org/10.18653/v1/2020.emnlp-main.550} {Dense
  passage retrieval for open-domain question answering}.
\newblock In \emph{Proceedings of the 2020 Conference on Empirical Methods in
  Natural Language Processing (EMNLP)}, pages 6769--6781, Online. Association
  for Computational Linguistics.

\bibitem[{Lu et~al.(2021)Lu, He, Xiong, Ke, Malik, Dou, Bennett, Liu, and
  Overwijk}]{lu-etal-2021-less}
Shuqi Lu, Di~He, Chenyan Xiong, Guolin Ke, Waleed Malik, Zhicheng Dou, Paul
  Bennett, Tie-Yan Liu, and Arnold Overwijk. 2021.
\newblock \href {https://doi.org/10.18653/v1/2021.emnlp-main.220} {Less is
  more: Pretrain a strong {S}iamese encoder for dense text retrieval using a
  weak decoder}.
\newblock In \emph{Proceedings of the 2021 Conference on Empirical Methods in
  Natural Language Processing}, pages 2780--2791, Online and Punta Cana,
  Dominican Republic. Association for Computational Linguistics.

\bibitem[{Nguyen et~al.(2016)Nguyen, Rosenberg, Song, Gao, Tiwary, Majumder,
  and Deng}]{tri2016msmarco}
Tri Nguyen, Mir Rosenberg, Xia Song, Jianfeng Gao, Saurabh Tiwary, Rangan
  Majumder, and Li~Deng. 2016.
\newblock \href {https://ceur-ws.org/Vol-1773/CoCoNIPS\_2016\_paper9.pdf} {{MS}
  {MARCO:} {A} human generated machine reading comprehension dataset}.
\newblock In \emph{Proceedings of the Workshop on Cognitive Computation:
  Integrating neural and symbolic approaches 2016 co-located with the 30th
  Annual Conference on Neural Information Processing Systems {(NIPS} 2016),
  Barcelona, Spain, December 9, 2016}, volume 1773 of \emph{{CEUR} Workshop
  Proceedings}. CEUR-WS.org.

\bibitem[{Nogueira et~al.(2019)Nogueira, Yang, Lin, and
  Cho}]{Rodrigo2019doc2query}
Rodrigo~Frassetto Nogueira, Wei Yang, Jimmy Lin, and Kyunghyun Cho. 2019.
\newblock \href {http://arxiv.org/abs/1904.08375} {Document expansion by query
  prediction}.
\newblock \emph{CoRR}, abs/1904.08375.

\bibitem[{Qu et~al.(2021)Qu, Ding, Liu, Liu, Ren, Zhao, Dong, Wu, and
  Wang}]{qu-etal-2021-rocketqa}
Yingqi Qu, Yuchen Ding, Jing Liu, Kai Liu, Ruiyang Ren, Wayne~Xin Zhao, Daxiang
  Dong, Hua Wu, and Haifeng Wang. 2021.
\newblock \href {https://doi.org/10.18653/v1/2021.naacl-main.466}
  {{R}ocket{QA}: An optimized training approach to dense passage retrieval for
  open-domain question answering}.
\newblock In \emph{Proceedings of the 2021 Conference of the North American
  Chapter of the Association for Computational Linguistics: Human Language
  Technologies}, pages 5835--5847, Online. Association for Computational
  Linguistics.

\bibitem[{Robertson et~al.(2009)Robertson, Zaragoza
  et~al.}]{robertson2009probabilistic}
Stephen Robertson, Hugo Zaragoza, et~al. 2009.
\newblock The probabilistic relevance framework: Bm25 and beyond.
\newblock \emph{Foundations and Trends{\textregistered} in Information
  Retrieval}, 3(4):333--389.

\bibitem[{Shen et~al.(2022)Shen, Geng, Tao, Xu, Huang, Jiao, Yang, and
  Jiang}]{Tao2022lexmae}
Tao Shen, Xiubo Geng, Chongyang Tao, Can Xu, Xiaolong Huang, Binxing Jiao,
  Linjun Yang, and Daxin Jiang. 2022.
\newblock \href {https://doi.org/10.48550/arXiv.2208.14754} {Lexmae:
  Lexicon-bottlenecked pretraining for large-scale retrieval}.
\newblock \emph{CoRR}, abs/2208.14754.

\bibitem[{Thakur et~al.(2021)Thakur, Reimers, R{\"{u}}ckl{\'{e}}, Srivastava,
  and Gurevych}]{thakur2021beir}
Nandan Thakur, Nils Reimers, Andreas R{\"{u}}ckl{\'{e}}, Abhishek Srivastava,
  and Iryna Gurevych. 2021.
\newblock \href {http://arxiv.org/abs/2104.08663} {{BEIR:} {A} heterogenous
  benchmark for zero-shot evaluation of information retrieval models}.
\newblock \emph{CoRR}, abs/2104.08663.

\bibitem[{Wang et~al.(2022{\natexlab{a}})Wang, Yang, Huang, Jiao, Yang, Jiang,
  Majumder, and Wei}]{wang2022simlm}
Liang Wang, Nan Yang, Xiaolong Huang, Binxing Jiao, Linjun Yang, Daxin Jiang,
  Rangan Majumder, and Furu Wei. 2022{\natexlab{a}}.
\newblock \href {https://doi.org/10.48550/arXiv.2207.02578} {Simlm:
  Pre-training with representation bottleneck for dense passage retrieval}.
\newblock \emph{CoRR}, abs/2207.02578.

\bibitem[{Wang et~al.(2022{\natexlab{b}})Wang, Bao, Dong, Bjorck, Peng, Liu,
  Aggarwal, Mohammed, Singhal, Som, and Wei}]{wang2022beit3}
Wenhui Wang, Hangbo Bao, Li~Dong, Johan Bjorck, Zhiliang Peng, Qiang Liu, Kriti
  Aggarwal, Owais~Khan Mohammed, Saksham Singhal, Subhojit Som, and Furu Wei.
  2022{\natexlab{b}}.
\newblock \href {https://doi.org/10.48550/arXiv.2208.10442} {Image as a foreign
  language: Beit pretraining for all vision and vision-language tasks}.
\newblock \emph{CoRR}, abs/2208.10442.

\bibitem[{Wang et~al.(2021)Wang, Bao, Dong, and Wei}]{Wenhui2021VLMo}
Wenhui Wang, Hangbo Bao, Li~Dong, and Furu Wei. 2021.
\newblock \href {http://arxiv.org/abs/2111.02358} {Vlmo: Unified
  vision-language pre-training with mixture-of-modality-experts}.
\newblock \emph{CoRR}, abs/2111.02358.

\bibitem[{Wu et~al.(2022{\natexlab{a}})Wu, Ma, and Hu}]{wu2022query-as-context}
Xing Wu, Guangyuan Ma, and Songlin Hu. 2022{\natexlab{a}}.
\newblock \href {https://doi.org/10.48550/arXiv.2212.09598} {Query-as-context
  pre-training for dense passage retrieval}.
\newblock \emph{CoRR}, abs/2212.09598.

\bibitem[{Wu et~al.(2022{\natexlab{b}})Wu, Ma, Lin, Lin, Wang, and
  Hu}]{wu2022contextual}
Xing Wu, Guangyuan Ma, Meng Lin, Zijia Lin, Zhongyuan Wang, and Songlin Hu.
  2022{\natexlab{b}}.
\newblock \href {https://doi.org/10.48550/ARXIV.2208.07670} {Contextual masked
  auto-encoder for dense passage retrieval}.

\bibitem[{Wu et~al.(2023)Wu, Ma, Wang, Lin, Lin, Zhang, and
  Hu}]{wu2023cotmaev2}
Xing Wu, Guangyuan Ma, Peng Wang, Meng Lin, Zijia Lin, Fuzheng Zhang, and
  Songlin Hu. 2023.
\newblock \href {http://arxiv.org/abs/2304.03158} {Cot-mae v2: Contextual
  masked auto-encoder with multi-view modeling for passage retrieval}.

\bibitem[{Xiao et~al.(2022)Xiao, Liu, Shao, and Cao}]{xiao-etal-2022-retromae}
Shitao Xiao, Zheng Liu, Yingxia Shao, and Zhao Cao. 2022.
\newblock \href {https://aclanthology.org/2022.emnlp-main.35} {{R}etro{MAE}:
  Pre-training retrieval-oriented language models via masked auto-encoder}.
\newblock In \emph{Proceedings of the 2022 Conference on Empirical Methods in
  Natural Language Processing}, pages 538--548, Abu Dhabi, United Arab
  Emirates. Association for Computational Linguistics.

\bibitem[{Zhang et~al.(2022)Zhang, Tao, Shen, Xu, Geng, Jiao, and
  Jiang}]{zhang2022LED}
Kai Zhang, Chongyang Tao, Tao Shen, Can Xu, Xiubo Geng, Binxing Jiao, and Daxin
  Jiang. 2022.
\newblock \href {https://doi.org/10.48550/arXiv.2208.13661} {{LED:}
  lexicon-enlightened dense retriever for large-scale retrieval}.
\newblock \emph{CoRR}, abs/2208.13661.

\bibitem[{Zhou et~al.(2022)Zhou, Liu, Gong, Zhao, Jiang, Duan, and
  Wen}]{zhou2022master}
Kun Zhou, Xiao Liu, Yeyun Gong, Wayne~Xin Zhao, Daxin Jiang, Nan Duan, and
  Ji-Rong Wen. 2022.
\newblock Master: Multi-task pre-trained bottlenecked masked autoencoders are
  better dense retrievers.
\newblock \emph{arXiv preprint arXiv:2212.07841}.

\end{thebibliography}
\bibliographystyle{acl_natbib}

\appendix

\section{Appendix}
\label{sec:appendix}

\subsection{KL and JS Divergence}
\label{appendix_kl_js_div}
The distance matrix $d$ of queries $q$ and passages $p$ is calculated as follows.

\begin{equation}
d = 1 - s(q, p)
\end{equation}

\noindent where $s(q, p)$ is the cosine similarities of queries and passages. Then we calculate the Gaussian probability distribution (p) of each point of q-p pairs in two-dimensional spaces with Kernel Density Estimation.

\begin{equation}
p_{1} = \exp -d / \sum \exp -d
\end{equation}

\begin{equation}
p_{2} = \exp -d^T / \sum \exp -d^T 
\end{equation}

KL divergence is an asymmetric relative entropy. Here we calculate KL divergence based on query-to-passage distances for evaluating their mutual information.

\begin{equation}
KL = E(p_1, p_2) = \sum (p_1 * \log(p_1/p_2))
\end{equation}

\noindent where $E$ means entropy calculation. 

JS divergence is a symmetric version of KL divergence that measures the similarity between query-passage probability distributions.

\begin{equation}
p_{avg} = (p1+p2)/2
\end{equation}

\begin{equation}
JS = (E(p1, p_{avg}) + E(p2, p_{avg}))/2
\end{equation}

\subsection{In-domain Evaluation}
In Table \ref{table_appendix_score_s1_s2}, the results of CoT-MoTE on MS-MARCO passage ranking benchmark with BM25 and retriever-mined negatives are presented as follows.

\begin{table}[h]
\centering
\resizebox{\linewidth}{!}{
    \begin{tabular}{l|cccc}
    \toprule  
    CoT-MoTE & MRR@10 & R@1 & R@50 & R@1k \\
    \midrule
    \multicolumn{5}{l}{\textbf{BM25 negatives}} \\
    \midrule
    Dense & 37.7 & 24.9 & 86.8 & 98.4 \\
    Sparse & 39.6 & 26.7 & 87.6 & 98.7 \\
    Hybrid & 40.5 & 27.3 & 88.7 & 98.4 \\

    \midrule
    \multicolumn{5}{l}{\textbf{Mined negatives}} \\
    \midrule
    Dense & 39.7 & 26.7 & 88.1 & 98.7 \\
    Sparse & 40.5 & 27.1 & 88.4 & 98.8 \\
    Hybrid & 42.0 & 28.6 & 89.7 & 98.7 \\

    \bottomrule
    \end{tabular}
}
\caption{Ablation study for dense, sparse, and hybrid retrieval performances using BM25 negative and retriever-s1 mined negatives of CoT-MoTE. 
}
\label{table_appendix_score_s1_s2}
\end{table}

\begin{table*}[ht]
\centering
\resizebox{\linewidth}{!}{
    \begin{tabular}{l|c|c|c|c|c|c|c|c|c}
    \toprule  
    \textbf{Dataset} & BM25 & DPR & ANCE & ColBERT & Contriever & RetroMAE & MASTER  & CoT-MAE v2 & CoT-MoTE \\
     \midrule
    TREC-COVID & 65.6  & 33.2  & 65.4  & 67.7  & 59.6  & 75.6 & 62.0 & \textbf{77.1} & 75.2  \\ 
    NFCorpus & 32.5  & 18.9  & 23.7  & 30.5  & 32.8  & 30.1 & 33.0 & \textbf{33.5} & 33.1 \\ 
     \midrule
    NQ & 32.9  & 47.4  & 44.6  & 52.4  & 49.8  & 49.0  & 51.6 & 53.9 & \textbf{57.3} \\ 
    HotpotQA & 60.3  & 39.1  & 45.6  & 59.3  & 63.8  & 63.8  & 58.9 & \textbf{67.2} & 66.8  \\ 
    FiQA-2018 & 23.6  & 11.2  & 29.5  & 31.7  & 32.9  & 30.1 & 32.8 & 33.1 & \textbf{35.3} \\ 
     \midrule
    ArguAna & 31.5  & 17.5  & 41.5  & 23.3  & 44.6  & 48.1 & 39.5 & 48.2 & \textbf{50.2}  \\ 
    Touché-2020 & \textbf{36.7}  & 13.1  & 24.0  & 20.2  & 23.0  & 24.3 & 32.0 & 30.3 & 29.0 \\ 
     \midrule
    CQADupStack & 29.9  & 15.3  & 29.6  & 35.0  & 34.5  & \textbf{38.2} & 32.7 & 32.2 & 34.8 \\ 
    Quora & 78.9  & 24.8  & 85.2  & 85.4  & \textbf{86.5}  & 85.6 & 79.1  & 86.1 & 75.4 \\ 
     \midrule
    DBPedia & 31.3  & 26.3  & 28.1  & 39.2  & 41.3  & 38.5  & 39.9 & \textbf{42.6} & 42.0 \\ 
     \midrule
    SCIDOCS & 15.8  & 7.7  & 12.2  & 14.5  & \textbf{16.5}  & 15.0  & 14.1 & 16.5 & \textbf{16.7}  \\ 
     \midrule
    FEVER & 75.3  & 56.2  & 66.9  & 77.1  & 75.8  & 71.9  & 69.2 & \textbf{81.2} & 74.1 \\ 
    Climate-FEVER & 21.3  & 14.8  & 19.8  & 18.4  & 23.7  & 21.4  & 21.5 & \textbf{27.5} & 24.8  \\ 
    SciFact & 66.5  & 31.8  & 50.7  & 67.1  & 67.7  & 64.8  & 63.7 & 69.2 & \textbf{71.0}  \\ 
     \midrule
    Average & 43.0  & 25.5  & 40.5  & 44.4  & 46.6  & 46.9  & 45.0 & \textbf{49.9} & 49.0 \\ 
    \bottomrule
    \end{tabular}
}
\caption{Out-of-domain evaluation on BEIR benchmark. The score that is better in comparison is marked in bold.}
\label{table_beir_evaluation}
\end{table*}

\subsection{Out-of-domain Evaluation}
We also follow the CoT-MoE and try to pre-train our MoTE on Wiki corpus. Then we fine-tune with the MS-MARCO negatives and evaluate on BEIR benchmark \cite{thakur2021beir}. CoT-MoTE achieves better results on NQ, FiQA-2018, ArguAna, SCIDOCS, and SciFact. But it still slightly underperforms CoT-MoE v2. We guess the multi-view pre-training introduced by CoT-MoE v2 is more suitable for out-of-domain evaluation. we will keep working on investigating this phenomenon.

\end{document}